# MoFA: Model-based Deep Convolutional Face Autoencoder for Unsupervised Monocular Reconstruction

Ayush Tewari[1]   Michael Zollhöfer[1]   Hyeongwoo Kim[1]   Pablo Garrido[1]
Florian Bernard[1,2]   Patrick Pérez[3]   Christian Theobalt[1]
[1]Max-Planck-Institute for Informatics   [2] LCSB, University of Luxembourg   [3]Technicolor

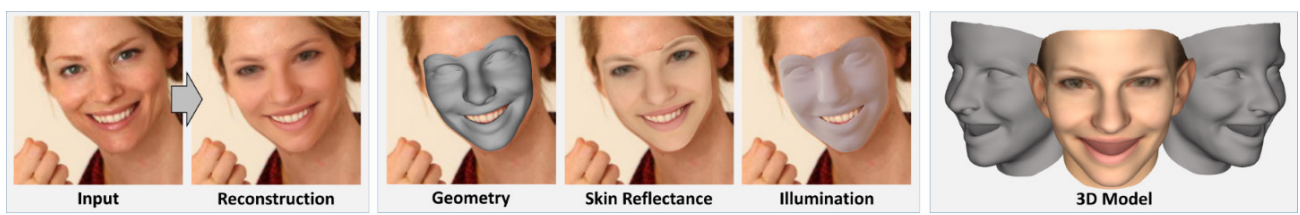

Our model-based deep convolutional face autoencoder enables unsupervised learning of semantic pose, shape, expression, reflectance and lighting parameters. The trained encoder predicts these parameters from a single monocular image, all at once.

## Abstract

*In this work we propose a novel model-based deep convolutional autoencoder that addresses the highly challenging problem of reconstructing a 3D human face from a single in-the-wild color image. To this end, we combine a convolutional encoder network with an expert-designed generative model that serves as decoder. The core innovation is the differentiable parametric decoder that encapsulates image formation analytically based on a generative model. Our decoder takes as input a code vector with exactly defined semantic meaning that encodes detailed face pose, shape, expression, skin reflectance and scene illumination. Due to this new way of combining CNN-based with model-based face reconstruction, the CNN-based encoder learns to extract semantically meaningful parameters from a single monocular input image. For the first time, a CNN encoder and an expert-designed generative model can be trained end-to-end in an unsupervised manner, which renders training on very large (unlabeled) real world data feasible. The obtained reconstructions compare favorably to current state-of-the-art approaches in terms of quality and richness of representation.*

## 1. Introduction

Detailed, dense 3D reconstruction of the human face from image data is a longstanding problem in computer vision and computer graphics. Previous approaches have tackled this challenging problem using calibrated multi-view data or uncalibrated photo collections [26, 45]. Robust and detailed three-dimensional face reconstruction from a single arbitrary in-the-wild image, e.g., downloaded from the Internet, is still an open research problem due to the high degree of variability of uncalibrated photos in terms of resolution and employed imaging device. In addition, in unconstrained photos, faces show a high variability in global pose, facial expression, and are captured under diverse and difficult lighting. Detailed 3D face reconstruction is the foundation for a broad scope of applications, which range from robust face recognition, over emotion estimation, to complex image manipulation tasks. In many applications, faces should ideally be reconstructed in terms of meaningful low-dimensional model parameters, which facilitates interpretation and manipulation of reconstructions (cf. [54]).

Recent monocular reconstruction methods broadly fall into two categories: Generative and regression-based. Generative approaches fit a parametric face model to image and video data, e.g., [3, 2, 12] by optimizing the alignment between the projected model and the image [14, 54, 51, 52, 25]. State-of-the-art generative approaches capture very detailed and complete 3D face models on the basis of semantically meaningful low-dimensional parameterizations [14, 54]. Unfortunately, the fitting energies are usually highly non-convex. Good results thus require an initialization close to the global optimum, which is only possible with some level of control during image capture or additional input data, e.g., detected landmarks.

Only recently, first regression-based approaches for dense 3D face reconstruction based on deep convolutional neural networks were proposed. Richardson et al. [41] use iterative regression to obtain a high quality estimate of pose, shape and expression, and finer scale surface detail [42] of a face model. The expression-invariant regression approach of Tran et al. [55] obtains high-quality estimates of shape and skin reflectance. Unfortunately, these approaches can only be trained in a supervised fashion on corpora of densely annotated facial images, the creation of which is a major obstacle in practice. In particular, the creation of a training corpus of photo-realistic synthetic facial images that include facial hair, parts of the upper body and a consistent background is challenging. While the refinement network of Richardson et al. [42] can be trained in an unsupervised manner, their coarse shape regression network requires synthetic ground truth data for training. Also, the quality and richness of representation (e.g. illumination and colored reflectance in addition to geometry) of these methods does not match the best generative ones. However, trained networks are efficient to evaluate and can be trained to achieve remarkable robustness under difficult real world conditions.

This paper contributes a new type of model-based deep convolutional autoencoder that joins forces of state-of-the-art generative and CNN-based regression approaches for dense 3D face reconstruction via a deep integration of the two on an architectural level. Our network architecture is inspired by recent progress on deep convolutional autoencoders, which, in their original form, couple a CNN encoder and a CNN decoder through a code-layer of reduced dimensionality [18, 33, 61]. Unlike previously used CNN-based decoders, our convolutional autoencoder deeply integrates an expert-designed decoder. This layer implements, in closed form, an elaborate generative analytically-differentiable image formation model on the basis of a detailed parametric 3D face model [3]. Some previous fully CNN-based autoencoders tried to disentangle [28, 15], but could not fully guarantee the semantic meaning of code layer parameters. In our new network, exact semantic meaning of the code vector, i.e., the input to the decoder, is ensured by design. Moreover, our decoder is compact and does not need training of enormous sets of unintuitive CNN weights.

Unlike previous CNN regression-based approaches for face reconstruction, a single forward pass of our network estimates a much more complete face model, including pose, shape, expression, skin reflectance, and illumination, at a high quality. Our new network architecture allows, for the first time, combined end-to-end training of a sophisticated model-based (generative) decoder and a CNN encoder, with error backpropagation through all layers. It also allows, for the first time, unsupervised training of a network that reconstructs dense and semantically meaningful faces on unlabeled in-the-wild images via a dense photometric training loss. In consequence, our network generalizes better to real world data compared to networks trained on synthetic face data [41, 42].

## 2. Related Work

In this section, we summarize previous works that are related to our approach. We focus on parametric model fitting and CNN approaches in the context of monocular face reconstruction. For further work on general template-based mesh tracking, please refer to [25, 26, 51, 52, 31].

**Parametric Face Models** Active Appearance Models (AAMs) use a linear model for jointly capturing shape and texture variation in 2D [9]. Matching an AAM to an image is a registration problem, usually tackled via energy optimization. A closely related approach to AAMs is the 3D morphable model of faces (3DMM) [3], which has been used for learning facial animations from 3D scans [2]. In [12], a parametric head model has been employed to modify the relative head pose and camera parameters of portrait photos.

**Monocular Optimization-based Reconstruction** Many monocular reconstruction approaches solve an optimization problem to fit a model to a given image. For example, the 3DMM has been used for monocular reconstruction [43] and image collection-based reconstruction [45]. In [14], high-quality 3D face rigs are obtained from monocular RGB video based on a multi-layer model. Even real-time facial reconstruction and reenactment has been achieved [54, 20]. Compared to optimization-based approaches, ours differs in two main regards. First, our network efficiently *regresses model parameters* without requiring iterative optimization. Second, given a cropped face image, our method does *not require an initialization* of the model parameters, which is a significant advantage over optimization-based techniques.

**Deep Learning for Coarse Face Reconstruction** The detection of facial landmarks in images is an active area of research [57, 23]. Various approaches are based on deep learning, including convolutional neural network (CNN) cascades [50, 63], a deep face shape model based on Restricted Boltzmann Machines [58], a recurrent network with long-short term memory [61], a recurrent encoder-decoder network for real-time facial landmark detection in video [38], or a two-stage convolutional part heatmap regression approach [5]. In [40], a multi-task CNN is trained to predict several face-related parameters (e.g. pose, gender, age), in addition to facial landmarks. These deep learning approaches share common limitations: They are trained in a *supervised* manner and *predict only sparse information*. In contrast, our approach works *unsupervised* and obtains a *dense* reconstruction via regressing generative model parameters.

**Deep Learning for Dense Face Reconstruction** Apart from the approaches mentioned above, there exist several dense deep learning approaches. A multilayer generative model based on deep belief networks for the generation of images under new lighting has been introduced in [53]. The face identity-preserving (FIP) descriptor has been proposed for reconstructing a face image in a canonical view [65]. The *Multi-View Perceptron* approach for face recognition learns disentangled view and facial identity parameters based on a training corpus that provides annotations of these dimensions [66]. The generation of faces from attributes [30] and dense shape regression [16] have also been studied. Non-linear variants of AAMs based on Deep Boltzmann Machines have been presented in [11, 39]. In [41], a CNN is trained using synthetic data for extracting the face geometry from a single image. Unsupervised refinement of these reconstructions has been proposed in [42]. [55] used photo collections to obtain the ground truth parameters from which a CNN is trained for regressing facial identity. In [29], a CNN is trained under controlled conditions in a supervised fashion for facial animation tasks. A framework for face hallucination from low-resolution face images has been proposed in [64]. All the discussed approaches require annotated training data. Since the annotation of a large image body is extremely expensive, some approaches (e.g. [41, 42]) resort to synthetic data. However, synthetic renderings usually lack realistic features, which has a negative impact on the reconstruction accuracy. In contrast, our approach uses *real data* and does not require ground truth model parameters.

**Autoencoders** Autoencoders approximate the identity mapping by coupling an encoding stage with a decoding stage to learn a *compact intermediate description*, the so-called *code vector*. They have been used for nonlinear dimensionality reduction [18] and to extract biologically plausible image features [33]. An appealing characteristic is that these architectures are in general *unsupervised*, i.e., no labeled data is required. Closely related are approaches that consider the encoding or decoding stage individually, such as inverting a generative model [35], or generating images from code vectors [62]. Autoencoders have been used to tackle a wide range of face-related tasks, including stacked progressive autoencoders for face recognition [24], real-time face alignment [60], face recognition using a supervised autoencoder [13], learning of face representations with a stacked autoencoder [10], or face de-occlusion [61]. The *Deep Convolutional Inverse Graphics Network* (DC-IGN) learns *interpretable* graphics codes that allow the reproduction of images under different conditions (e.g. pose and lighting) [28]. This is achieved by using mini-batches where only a single scene parameter is known to vary. The disentanglement of code variables, such as shape and scene-related transformations has been considered in [15]. Our proposed approach stands out from existing techniques, since we consider the *full set of meaningful parameters* and do not need to group images according to known variations.

**Deep Integration of Expert Layers** Inspired by *Spatial Transformer Networks* [21], the *gvvn* library implements low-level geometric computer vision layers [17]. Unsupervised volumetric 3D object reconstruction from a single-view by *Perspective Transformer Nets* has been demonstrated in [59]. Unlike these approaches, we tackle a higher level computer vision task, namely the monocular reconstruction of semantically meaningful parameters for facial geometry, expression, illumination, and camera extrinsics.

## 3. Overview

Our novel deep convolutional model-based face autoencoder enables unsupervised end-to-end learning of meaningful semantic face and rendering parameters, see Fig. 1. To this end, we combine convolutional encoders with an expert-designed differentiable model-based decoder that analytically implements image formation. The decoder generates a realistic synthetic image of a face and enforces semantic meaning by design. Rendering is based on an image formation model that enforces full semantic meaning via a parametric face prior. More specifically, we independently parameterize pose, shape, expression, skin reflectance and illumination. The synthesized image is compared to the input image using a robust photometric loss $E_{\text{loss}}$ that includes statistical regularization of the face. In combination, this enables unsupervised end-to-end training of our networks. 2D facial landmark locations can be optionally provided to add a surrogate loss for faster convergence and improved reconstructions, see Sec. 6. Note, both scenarios require no supervision of the semantic parameters. After training, the encoder part of the network enables regression of a dense face model and illumination from a single monocular image, without requiring any other input, such as landmarks.

## 4. Semantic Code Vector

The semantic code vector $\mathbf{x} \in \mathbb{R}^{257}$ parameterizes the facial expression $\boldsymbol{\delta} \in \mathbb{R}^{64}$, shape $\boldsymbol{\alpha} \in \mathbb{R}^{80}$, skin reflectance $\boldsymbol{\beta} \in \mathbb{R}^{80}$, camera rotation $\mathbf{T} \in SO(3)$ and translation $\mathbf{t} \in \mathbb{R}^3$, and the scene illumination $\boldsymbol{\gamma} \in \mathbb{R}^{27}$ in a unified manner:

$$\mathbf{x} = (\underbrace{\boldsymbol{\alpha}, \boldsymbol{\delta}, \boldsymbol{\beta}}_{\text{face}}, \underbrace{\mathbf{T}, \mathbf{t}, \boldsymbol{\gamma}}_{\text{scene}}) \ . \quad (1)$$

In the following, we describe the parameters that are associated with the employed face model. The parameters that govern image formation are described later on in Sec. 5.

The face is represented as a manifold triangle mesh with $N = 24\text{k}$ vertices $\mathbf{V} = \{\mathbf{v}_i \in \mathbb{R}^3 | 1 \leq i \leq N\}$. The associated vertex normals $\mathbf{N} = \{\mathbf{n}_i \in \mathbb{R}^3 | 1 \leq i \leq N\}$ are computed using a local one-ring neighborhood. The spatial embedding $\mathbf{V}$ is parameterized by an affine face model:

$$\mathbf{V} = \hat{\mathbf{V}}(\boldsymbol{\alpha}, \boldsymbol{\delta}) = \mathbf{A}_s + \mathbf{E}_s \boldsymbol{\alpha} + \mathbf{E}_e \boldsymbol{\delta} \ . \quad (2)$$

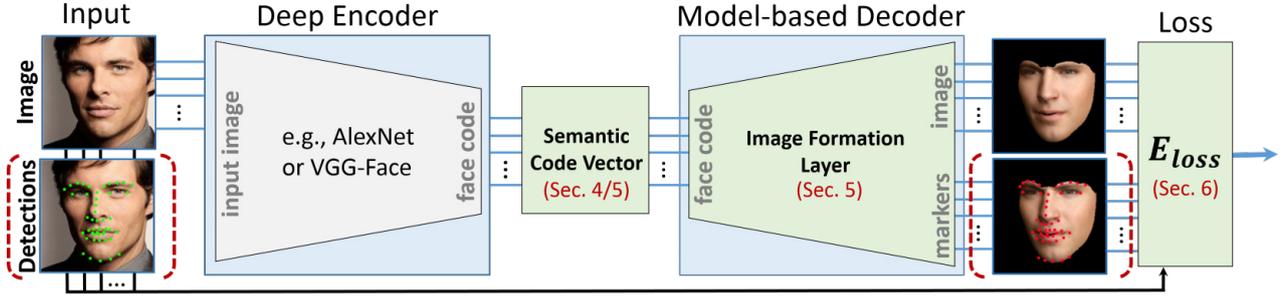

Figure 1. Our deep model-based face autoencoder enables unsupervised end-to-end learning of semantic parameters, such as pose, shape, expression, skin reflectance and illumination. An optional landmark-based surrogate loss enables faster convergence and improved reconstruction results, see Sec. 6. Both scenarios require no supervision of the semantic parameters during training.

Here, the average face shape $\mathbf{A}_s$ has been computed based on 200 (100 male, 100 female) high-quality face scans [3]. The linear PCA bases $\mathbf{E}_s \in \mathbb{R}^{3N \times 80}$ and $\mathbf{E}_e \in \mathbb{R}^{3N \times 64}$ encode the modes with the highest shape and expression variation, respectively. We obtain the expression basis by applying PCA to the combined set of blendshapes of [1] and [6], which have been re-targeted to the face topology of [3] using deformation transfer [49]. The PCA basis covers more than 99% of the variance of the original blendshapes.

In addition to facial geometry, we also parameterize per-vertex skin reflectance $\mathbf{R} = \{\mathbf{r}_i \in \mathbb{R}^3 | 1 \leq i \leq N\}$ based on an affine parametric model:

$$\mathbf{R} = \hat{\mathbf{R}}(\boldsymbol{\beta}) = \mathbf{A}_r + \mathbf{E}_r \boldsymbol{\beta} \ . \quad (3)$$

Here, the average skin reflectance $\mathbf{A}_r$ has been computed based on [3] and the orthogonal PCA basis $\mathbf{E}_r \in \mathbb{R}^{3N \times 80}$ captures the modes of highest variation. Note, all basis vectors are already scaled with the appropriate standard deviations $\sigma_k^\bullet$ such that $\mathbf{E}_\bullet^T \mathbf{E}_\bullet = \text{diag}(\cdots, [\sigma_k^\bullet]^2, \cdots)$.

## 5. Parametric Model-based Decoder

Given a scene description in the form of a semantic code vector $\mathbf{x}$, our parametric decoder generates a realistic synthetic image of the corresponding face. Since our image formation model is fully analytical and differentiable, we also implement an efficient backward pass that inverts image formation via standard backpropagation. This enables unsupervised end-to-end training of our network. In the following, we describe the used image formation model.

**Perspective Camera** We render realistic facial imagery using a pinhole camera model under a full perspective projection $\Pi : \mathbb{R}^3 \to \mathbb{R}^2$ that maps from camera space to screen space. The position and orientation of the camera in world space is given by a rigid transformation, which we parameterize based on a rotation $\mathbf{T} \in \mathbf{SO}(3)$ and a global translation $\mathbf{t} \in \mathbb{R}^3$. Hence, the functions $\Phi_{\mathbf{T},\mathbf{t}}(\mathbf{v}) = \mathbf{T}^{-1}(\mathbf{v} - \mathbf{t})$ and $\Pi \circ \Phi_{\mathbf{T},\mathbf{t}}(\mathbf{v})$ map an arbitrary point $\mathbf{v}$ from world to camera space and further to screen space.

**Illumination Model** We represent scene illumination using Spherical Harmonics (SH) [34]. Here, we assume distant low-frequency illumination and a purely *Lambertian* surface reflectance. Thus, we evaluate the radiosity at vertex $\mathbf{v}_i$ with surface normal $\mathbf{n}_i$ and skin reflectance $\mathbf{r}_i$ as follows:

$$C(\mathbf{r}_i, \mathbf{n}_i, \boldsymbol{\gamma}) = \mathbf{r}_i \cdot \sum_{b=1}^{B^2} \boldsymbol{\gamma}_b \mathbf{H}_b(\mathbf{n}_i) \ . \quad (4)$$

The $\mathbf{H}_b : \mathbb{R}^3 \to \mathbb{R}$ are SH basis functions and the $B^2 = 9$ coefficients $\boldsymbol{\gamma}_b \in \mathbb{R}^3$ ($B = 3$ bands) parameterize colored illumination using the red, green and blue channel.

**Image Formation** We render realistic images of the scene using the presented camera and illumination model. To this end, in the forward pass $\mathcal{F}$, we compute the screen space position $\mathbf{u}_i(\mathbf{x})$ and associated pixel color $\mathbf{c}_i(\mathbf{x})$ for each vertex $\mathbf{v}_i$:

$$\begin{aligned}
\mathcal{F}_i(\mathbf{x}) &= [\mathbf{u}_i(\mathbf{x}), \mathbf{c}_i(\mathbf{x})]^T \in \mathbb{R}^5 \ , \quad (5)\\
\mathbf{u}_i(\mathbf{x}) &= \Pi \circ \Phi_{\mathbf{T},\mathbf{t}}(\hat{\mathbf{V}}_i(\boldsymbol{\alpha}, \boldsymbol{\delta})) \ , \\
\mathbf{c}_i(\mathbf{x}) &= C(\hat{\mathbf{R}}_i(\boldsymbol{\beta}), \mathbf{Tn}_i(\boldsymbol{\alpha}, \boldsymbol{\delta}), \boldsymbol{\gamma}) \ .
\end{aligned}$$

Here, $\mathbf{Tn}_i$ transforms the world space normals to camera space and $\gamma$ models illumination in camera space.

**Backpropagation** To enable training, we implement a backward pass that inverts image formation:

$$\mathcal{B}_i(\mathbf{x}) = \frac{\mathrm{d}\mathcal{F}_i(\mathbf{x})}{\mathrm{d}(\boldsymbol{\alpha}, \boldsymbol{\delta}, \boldsymbol{\beta}, \mathbf{T}, \mathbf{t}, \boldsymbol{\gamma})} \in \mathbb{R}^{5 \times 257} \ . \quad (6)$$

This requires the computation of the gradients of the image formation model (see Eq. 5) with respect to the face and scene parameters. For high efficiency during training, we evaluate the gradients in a data-parallel manner, see Sec. 7.

## 6. Loss Layer

We employ a robust dense photometric loss function that enables efficient end-to-end training of our networks. The loss is inspired by recent optimization-based approaches [14, 54] and combines three terms:

$$E_{\text{loss}}(\mathbf{x}) = \underbrace{w_{\text{land}} E_{\text{land}}(\mathbf{x}) + w_{\text{photo}} E_{\text{photo}}(\mathbf{x})}_{\text{data term}} + \underbrace{w_{\text{reg}} E_{\text{reg}}(\mathbf{x})}_{\text{regularizer}} . \quad (7)$$

Here, we enforce sparse landmark alignment $E_{\text{land}}$, dense photometric alignment $E_{\text{photo}}$ and statistical plausibility $E_{\text{reg}}$ of the modeled faces. Note, $E_{\text{land}}$ is optional and implements a surrogate loss that can be used to speed up convergence, see Sec. 7. The binary weight $w_{\text{land}} \in \{0, 1\}$ toggles this constraint. The constant weights $w_{\text{photo}} = 1.92$ and $w_{\text{reg}} = 2.9 \times 10^{-5}$ balance the contributions of the objectives.

**Dense Photometric Alignment** The goal of the encoder is to predict model parameters that lead to a synthetic face image that matches the provided monocular input image. To this end, we employ dense photometric alignment, similar to [54], on a per-vertex level using a robust $\ell_{2,1}$-norm:

$$E_{\text{photo}}(\mathbf{x}) = \frac{1}{N} \sum_{i \in \mathcal{V}} \left\| \mathcal{I}(\mathbf{u}_i(\mathbf{x})) - \mathbf{c}_i(\mathbf{x}) \right\|_2 . \quad (8)$$

Here, $\mathcal{I}$ is an image of the training corpus and we iterate over the set of front facing vertices $\mathcal{V}$, which we compute based on the current forward pass, for occlusion awareness.

**Sparse Landmark Alignment** In addition to dense photometric alignment, we propose an optional surrogate loss based on detected facial landmarks [46]. We use a subset of 46 landmarks (out of 66), see Fig. 1. Given the subset $\mathcal{L} = \{(\mathbf{s}_j, c_j, k_j)\}_{j=1}^{46}$ of detected 2D landmarks $\mathbf{s}_j \in \mathbb{R}^2$, with confidence $c_j \in [0, 1]$ (1 confident) and corresponding model vertex index $k_j \in \{1, ..., N\}$, we enforce the projected 3D vertices to be close to the 2D detections:

$$E_{\text{land}}(\mathbf{x}) = \sum_{j=1}^{46} c_j \cdot \left\| \mathbf{u}_{k_j}(\mathbf{x}) - \mathbf{s}_j \right\|_2^2 . \quad (9)$$

Please note, this surrogate loss is optional. Our networks can be trained fully unsupervised without supplying these sparse constraints. After training, landmarks are never required.

**Statistical Regularization** During training, we further constrain the optimization problem using statistical regularization [3] on the model parameters:

$$E_{\text{reg}}(\mathbf{x}) = \sum_{k=1}^{80} \boldsymbol{\alpha}_k^2 + w_{\boldsymbol{\beta}} \sum_{k=1}^{80} \boldsymbol{\beta}_k^2 + w_{\boldsymbol{\delta}} \sum_{k=1}^{64} \boldsymbol{\delta}_k^2 . \quad (10)$$

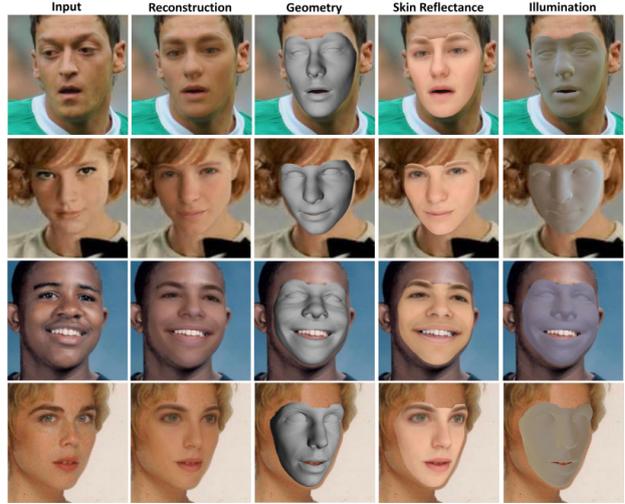

Figure 2. Our approach enables the regression of high quality pose, shape, expression, skin reflectance and illumination from just a single monocular image (images from CelebA [32]).

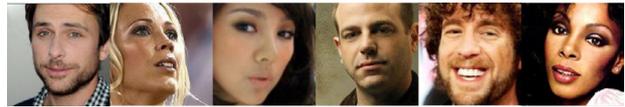

Figure 3. Sample images of our real world training corpus.

This constraint enforces plausible facial shape $\boldsymbol{\alpha}$, expression $\boldsymbol{\delta}$ and skin reflectance $\boldsymbol{\beta}$ by preferring values close to the average (the basis of the linear face model is already scaled by the standard deviations). The parameters $w_{\boldsymbol{\beta}} = 1.7 \times 10^{-3}$ and $w_{\boldsymbol{\delta}} = 0.8$ balance the importance of the terms. Note, we do not regularize pose $(\mathbf{T}, \mathbf{t})$ and illumination $\boldsymbol{\gamma}$.

**Backpropagation** To enable training based on stochastic gradient descent, during backpropagation, the gradient of the robust loss is passed backward to our model-based decoder and is combined with $\mathcal{B}_i(\mathbf{x})$ using the chain rule.

## 7. Results

We demonstrate unsupervised learning of our model-based autoencoder in the wild and also show that a surrogate loss during training improves accuracy. We test encoders based on AlexNet [27] and VGG-Face [37], where we modified the last fully connected layer to output our 257 model parameters. The reported results have been obtained using AlexNet [27] as encoder and without the surrogate loss, unless stated otherwise. After training, the encoder regresses pose, shape, expression, skin reflectance and illumination at once from a single image, see Fig. 2. For training we use an image corpus (see Fig. 3), which is a combination of four datasets: CelebA [32], LFW [19], Facewarehouse [7], and 300-VW [8, 48, 56]. The corpus is automatically annotated using facial landmark detection (see Sec. 6) and cropped to a bounding box using Haar Cascade Face Detection [4]. We prune frames with bad detections. The crops are scaled to a resolution of $240 \times 240$ pixels. In total, we collect 147k images, which we randomize and split into 142k for training and 5k for evaluation. We train our network using the *Caffe* [22] deep learning framework. For efficiency, we implement our model-based decoder and the robust photometric loss in a single CUDA [36] layer. We train our networks using *AdaDelta* and perform 200k batch iterations (batch size of 5). The base learning rate is $0.1$ for all parameters, except for the Z-translation, for which we set it to $0.0005$. At test time, regressing all parameters using a TitanX Pascal graphics card is fast and takes only 4ms (AlexNet) or 14ms (VGG-Face). Training takes 13 hours (AlexNet) or 20 hours (VGG-Face). The encoder is initialized based on the provided pre-trained weights. All weights in the last fully connected layer are initialized to zero. This guarantees that the initial prediction is the average face placed in the middle of the screen and lit by ambient light, which is a good initialization. Note, the ambient coefficients of our renderer have an offset of $0.7$ to guarantee that the scene is initially lit.

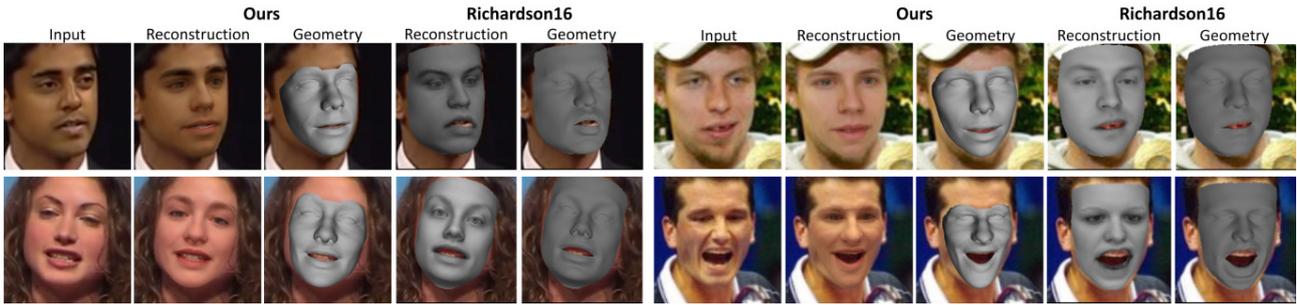

Figure 4. Comparison to Richardson et al. [41, 42] (coarse network without refinement) on 300-VW [8, 48, 56] (left) and LFW [19] (right). Our approach obtains higher reconstruction quality and provides estimates of colored reflectance and illumination. Note, the greyscale reflectance of [41, 42] is not regressed, but obtained via optimization, we on the other hand regress all parameters at once.

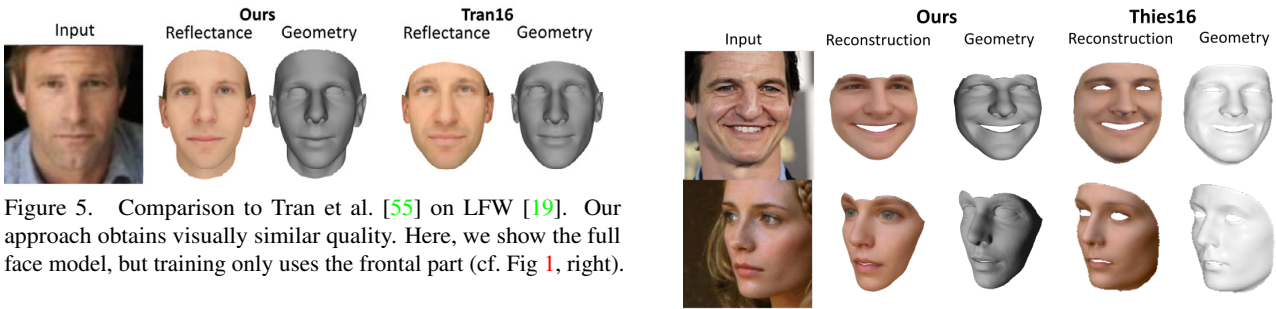

Figure 5. Comparison to Tran et al. [55] on LFW [19]. Our approach obtains visually similar quality. Here, we show the full face model, but training only uses the frontal part (cf. Fig 1, right).

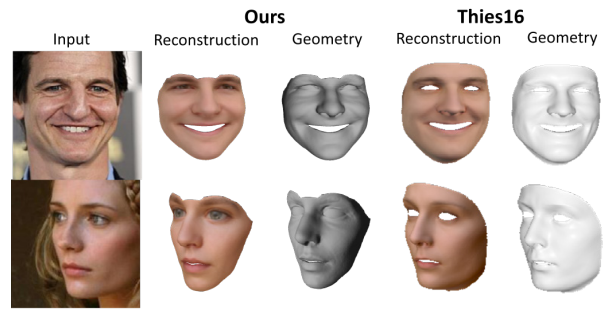

Figure 6. Comparison to the monocular reconstruction approach of [54] on CelebA [32]. Our approach obtains similar or higher quality, while being orders of magnitude faster (4ms vs. ~ 500ms).

Next, we compare to state-of-the-art optimization- and learning-based monocular reconstruction approaches, and evaluate all components of our approach.

**Comparison to Richardson et al. [41, 42]** We compare our approach to the CNN-based iterative regressor of Richardson et al. [41, 42]. Our results are compared qualitatively (Fig. 4) and quantitatively (Fig. 11) to their coarse regression network. Note, the refinement layer of [42] is orthogonal to our approach. Unlike [41, 42], our network is trained completely unsupervised on real images, while they use a synthetic training corpus that lacks realistic features. In contrast to [41, 42], we also regress colored skin reflectance and illumination, which is critical for many applications, e.g., relighting. Note, the greyscale reflectance of [41, 42] is not regressed, but obtained via optimization.

**Comparison to Tran et al. [55]** We compare qualitatively (Fig. 5) and quantitatively (Fig. 11) to the CNN-based identity regression approach of Tran et al. [55]. Our reconstructions are of visually similar quality, however, we additionally obtain high quality estimates of the facial expression and illumination. We also performed a face verification test on LFW. Our approach obtains an accuracy of 77%, which is higher than the monocular 3DMM baseline [44] (75%). Tran et al. [55] report an accuracy of 92%. Our approach is not designed for this scenario, since it is trained unsupervised on in-the-wild images. Tran et al. [55] require more supervision (photo collection) to train their network.

**Comparison to Thies et al. [54]** We compare our approach qualitatively (Fig. 6) and quantitatively (Fig. 11) to the state-of-the-art optimization-based monocular reconstruction approach of Thies et al. [54]. Our approach obtains similar or even higher quality, while being orders of magni-tude faster (4ms vs. ∼ 500ms). Note, while [54] tracks at real-time after identity estimation, it requires half a second to fit all parameters starting from the average model. While our approach only requires face detection at test time, Thies et al. [54] require detected landmarks.

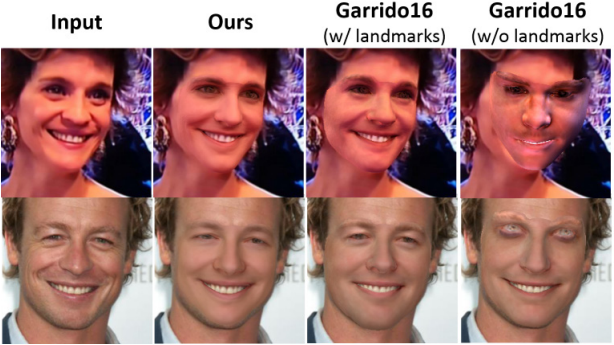

Figure 7. We compare to our implementation of the high quality off-line monocular reconstruction approach of [14]. We obtain similar quality without requiring landmarks as input. Without landmarks, [14] often gets stuck in a local minimum.

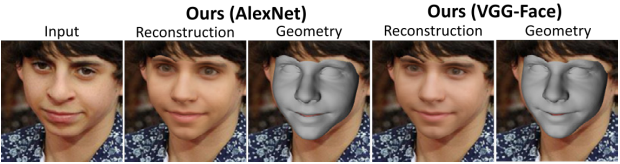

Figure 8. We evaluate different encoders in combination with our model-based decoder. In average VGG-Face [37] leads to slightly better results than AlexNet [27], but the results are comparable.

**Comparison to Garrido et al. [14]** We compare to our implementation (coarse layer, photometric + landmark + regularization terms, 50 Gauss-Newton steps) of the high quality off-line monocular reconstruction approach of [14], which requires landmarks as input. Our approach obtains comparable quality, while requiring no landmarks, see Fig. 7. Without sparse constraints as input, optimization-based approaches often get stuck in a local minimum.

**Evaluation of Different Encoders** We evaluate the impact of different encoders. VGG-Face [37] leads to slightly better results than AlexNet [27], see Fig. 8. On average VGG-Face [37] has a slightly lower landmark (4.9 pixels vs. 5.3 pixels) and photometric error (0.073 vs. 0.075, color distance in RGB space, each channel in [0, 1]), see Fig. 9.

**Quantitative Evaluation of Unsupervised Training** Unsupervised training decreases the dense photometric and landmark error (on a validation set of 5k real images), even when landmark alignment is not part of the loss function, see Fig. 9. The landmark error is computed based on 46 detected landmarks [46]. Training with our surrogate loss improves landmark alignment (AlexNet: 3.7 pixels vs. 5.3 pixels, VGG-Face: 3.4 pixels vs. 4.9) and leads to a similar photometric error (AlexNet: 0.078 vs. 0.075, VGG-Face: 0.078 vs. 0.073, color distance in RGB space, each channel in [0, 1]). We also evaluate the influence of our landmark-based surrogate loss qualitatively, see Fig. 12. Training with landmarks helps to improve robustness to occlusions and the quality of the predicted expressions. Note that both scenarios do not require landmarks, at test time.

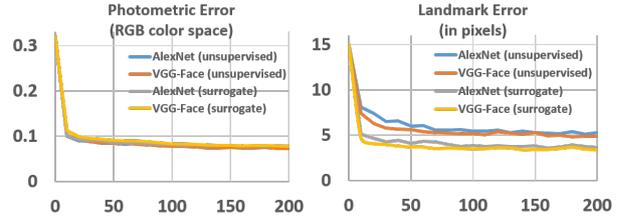

Figure 9. Quantitative evaluation on real data: Both landmark and photometric error are decreased during unsupervised training, even though landmark alignment is not part of the loss function.

Table 1. Quantitative evaluation on real data.

|  | Geometry | Photometric | Landmark |
|---|---|---|---|
| Ours (w/o surrogate) | 1.9mm | 0.065 | 5.0px |
| Ours (w/ surrogate) | 1.7mm | 0.068 | 3.2px |
| Garrido et al. [14] | 1.4mm | 0.052 | 2.6px |

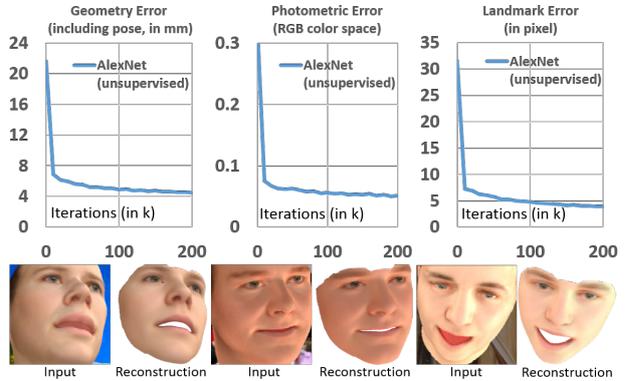

Figure 10. Quantitative evaluation on synthetic ground truth data: Training decreases the geometric, photometric and landmark error.

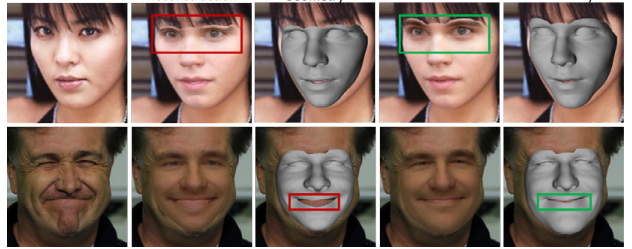

Figure 12. We evaluate the influence of the proposed surrogate task. The surrogate task leads to improved reconstruction quality and increases robustness to occlusions and strong expressions.

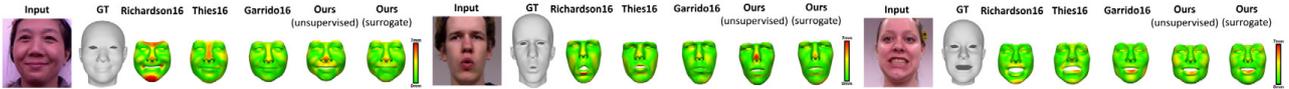

Figure 11. Quantitative evaluation on three images of Facewarehouse [7]: We obtain a low error that is comparable to optimization-based approaches. For this test, we trained our network using the intrinsics of the Kinect.

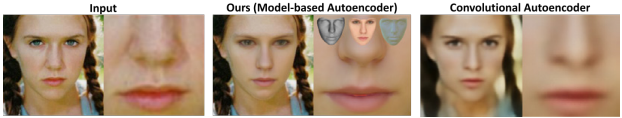

Figure 13. Our model-based autoencoder gives results of higher quality than convolutional autoencoders. In addition, it provides access to dense geometry, reflectance, and illumination.

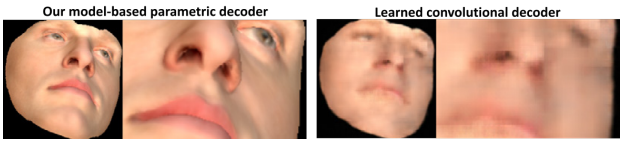

Figure 14. Our model-based decoder provides higher fidelity in terms of image quality than a learned convolutional decoder.

**Quantitative Evaluation** We perform a ground truth evaluation based on 5k rendered images with known parameters. Our model-based autoencoder (AlexNet, unsupervised) is trained on a corpus of 100k synthetic images with background augmentation, see Fig. 10. We measure the geometric error as the point-to-point 3D distance (including the estimated rotation, we compensate for translation and isotropic scale) between the estimate and the ground truth mesh. This error drops from 21.6mm to 4.5mm. The photometric error in RGB space also decreases (0.33 to 0.05) and so does the landmark error (31.6 pixels to 3.9 pixels). Overall, we obtain good fits. We also performed a quantitative comparison for 9 identities (180 images) on Facewarehouse, see Tab. 1 and Fig. 11. Our approach obtains low errors, which are on par with optimization-based techniques, while being much faster (4ms vs. 1min) and not requiring landmarks at test time.

**Comparison to Autoencoders and Learned Decoders**
We compare our model-based with a convolutional autoencoder in Fig. 13. The autoencoder uses four $3 \times 3$ convolution layers (64, 96, 128, 256 channels), a fully connected layer (257 outputs, same as number of model parameters), and four $4 \times 4$ deconvolution layers (128, 96, 64, 3 channels). Our model-based approach obtains sharper reconstruction results and provides fine granular semantic parameters allowing access to dense geometry, reflectance, and illumination, see Fig. 13 (middle). Explicit disentanglement [28, 15] of a convolutional autoencoder requires labeled ground truth data. We also compare to image formation based on a trained decoder. To this end, we train the decoder (similar parameters as above) based on synthetic imagery generated by our model to learn the parameter-to-image mapping. Our model-based decoder obtains renderings of higher fidelity compared to the learned decoder, see Fig. 14.

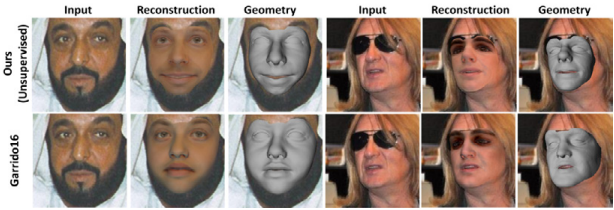

Figure 15. Facial hair and occlusions are challenging to handle.

## 8. Limitations

We have demonstrated compelling monocular reconstructions using a novel model-based autoencoder that is trained unsupervised. Similar to other regression approaches, implausible reconstructions are possible outside the span of training data. This can be alleviated by enlarging the training corpus, which is easy to achieve in our unsupervised setting. Since we employ a face model, reconstructions are limited to the modeled subspace. Similar to optimization-based approaches, strong occlusions, e.g., by facial hair or external objects, cause our approach to fail, see Fig. 15. Unsupervised occlusion-aware training is an interesting open research problem. Similar to related approaches, strong head rotations are challenging. Since we do not model the background, our reconstructions can slightly shrink. Shrinking is discussed and addressed in [47].

## 9. Conclusion

We have presented the first deep convolutional model-based face autoencoder that can be trained in an unsupervised manner and learns meaningful semantic parameters. Semantic meaning in the code vector is enforced by a parametric model that encodes variation along the pose, shape, expression, skin reflectance and illumination dimensions. Our model-based decoder is fully differentiable and allows end-to-end learning of our network.

We believe that the fundamental technical concepts of our approach go far beyond the context of monocular face reconstruction and will inspire future work.

## Acknowledgements

We thank True-VisionSolutions Pty Ltd for kindly providing the 2D face tracker, Anh Tuan Tran for publishing their source code, and Justus Thies and Elad Richardson for the comparisons. We thank Christian Richardt for the video voice over. This work was supported by the ERC Starting Grant CapReal (335545) and by Technicolor.

# MoFA: Model-based Deep Convolutional Face Autoencoder for Unsupervised Monocular Reconstruction
## — Supplemental Material —

Ayush Tewari[1]    Michael Zollhöfer[1]    Hyeongwoo Kim[1]    Pablo Garrido[1]
Florian Bernard[1,2]    Patrick Pérez[3]    Christian Theobalt[1]

[1]Max-Planck-Institute for Informatics    [2] LCSB, University of Luxembourg    [3]Technicolor

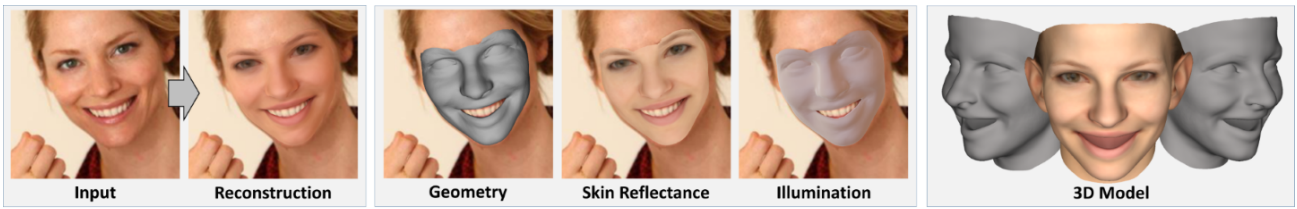

Our model-based deep convolutional face autoencoder enables unsupervised learning of semantic pose, shape, expression, reflectance and lighting parameters. The trained encoder predicts these parameters from a single monocular image, all at once.

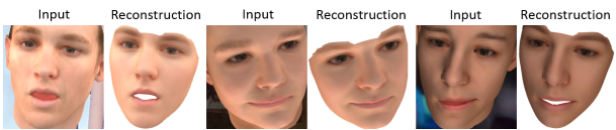

Figure 1. Results on synthetic ground truth data: We obtain good fits for all parameters.

This supplemental document shows more results and evaluations of our novel model-based deep convolutional face autoencoder (MoFA) that allows for unsupervised monocular reconstruction. In particular, we show more images of our real-world training corpus (see Fig. 3), additional qualitative results (see Fig. 4) and additional comparison to optimization-based (see Fig. 2 and 10) and learning based (see Fig. 9) monocular reconstruction approaches. We evaluate the influence of our surrogate task (see Fig. 7) and show reconstruction results based on different encoders (see Fig. 6). We provide a visual evaluation of the convergence (see Fig. 8). In addition, we illustrate the limitations of our approach (see Fig. 5) and show more reconstruction results on our synthetic ground truth test set (see Fig. 1). For a detailed description of these results, please refer to the corresponding sections of the main document.

Additional reconstruction results for images and video sequences are shown in the supplemental video. Note, to obtain temporally coherent video results we generate the 2D bounding box crops, which are the input to our network, using the face tracker of [10]. For all image results we obtain the crops using Haar Cascade Face Detection [1].

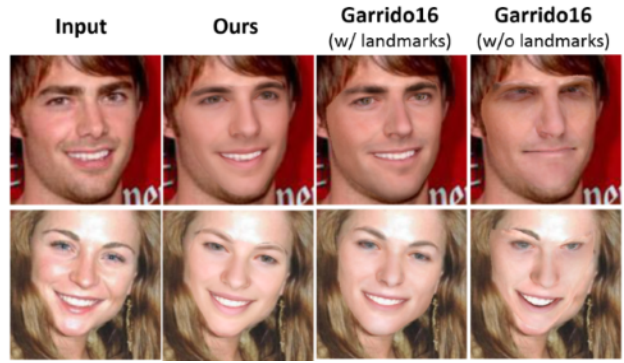

Figure 2. We compare to our implementation of the high quality off-line monocular reconstruction approach of [3]. We obtain similar quality without requiring landmarks as input. Without landmarks, [3] often gets stuck in a local minimum.

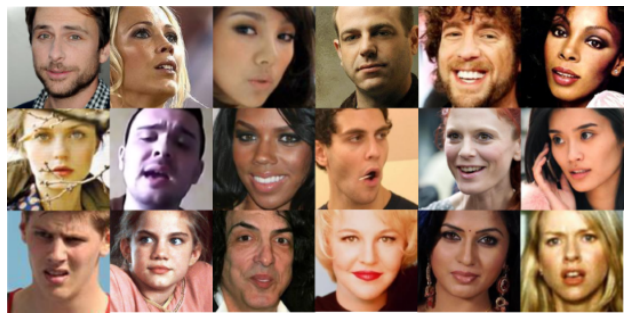

Figure 3. Sample images of our real world training corpus.

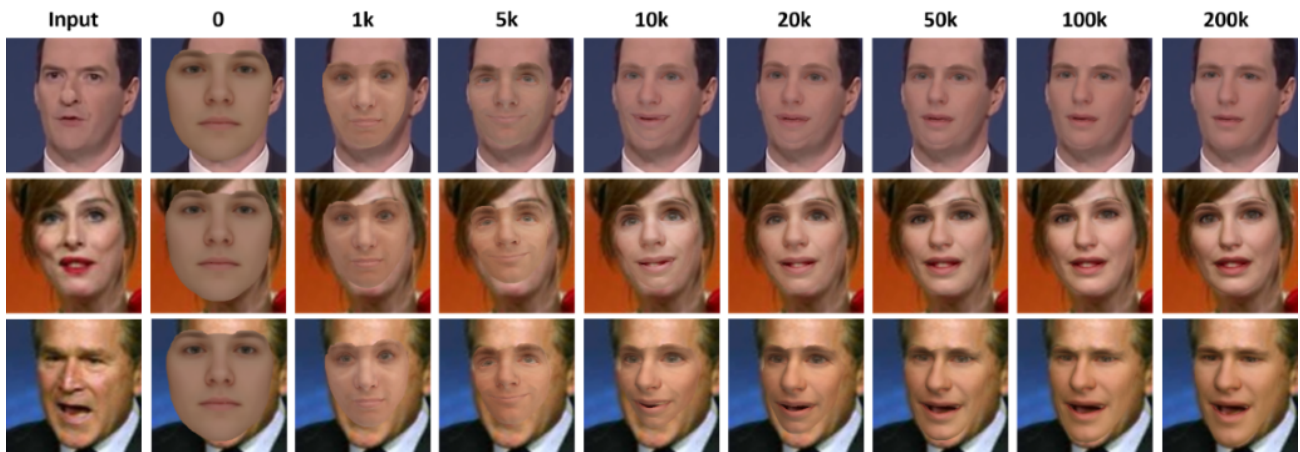

Figure 8. Visual evaluation of the convergence during training. Starting from the average face, our approach learns the variation between faces in an unsupervised manner. This evaluation has been performed on the test set — the CNN has not seen these images during training.

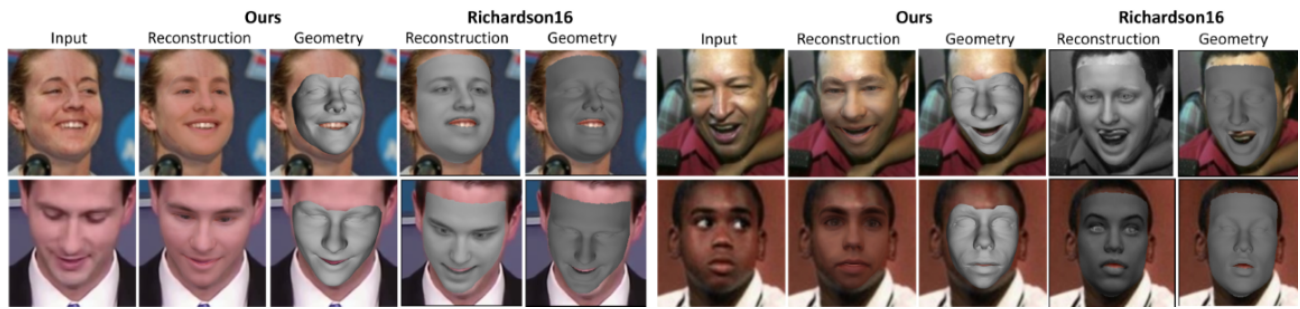

Figure 9. Comparison to Richardson et al. [8, 9] (coarse network without refinement) on 300-VW [2, 11, 13] (left) and LFW [4] (right). Our approach obtains higher reconstruction quality and provides estimates of colored reflectance and illumination. Note, the greyscale reflectance of [8, 9] is not regressed, but obtained via optimization, we on the other hand regress all parameters at once.

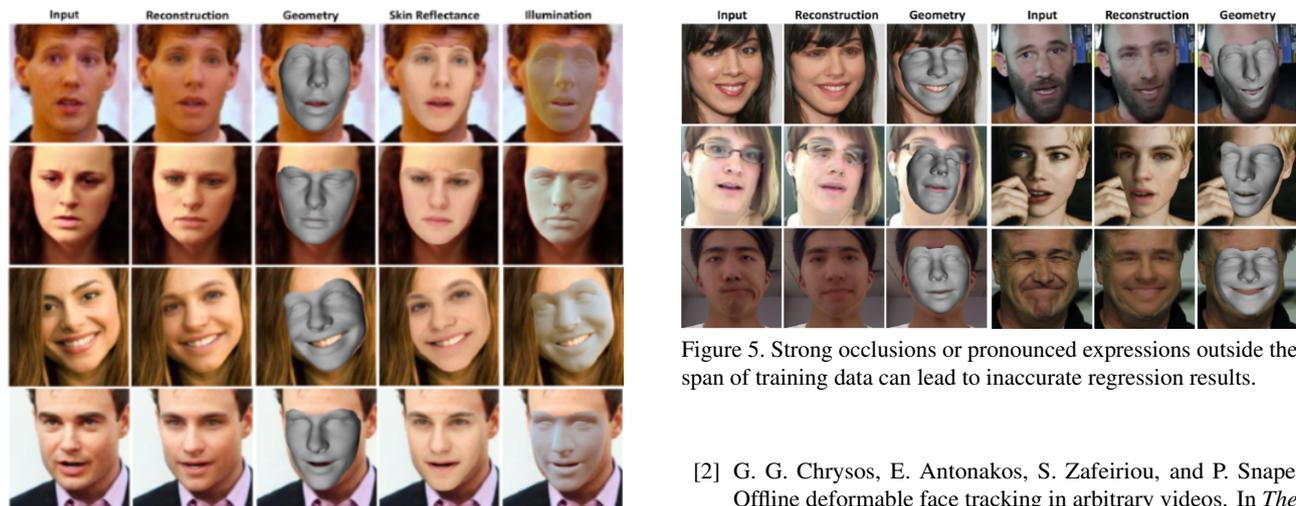

Figure 4. Our approach enables the regression of high quality pose, shape, expression, skin reflectance and illumination from just a single monocular image (images from CelebA [6]).

Figure 5. Strong occlusions or pronounced expressions outside the span of training data can lead to inaccurate regression results.

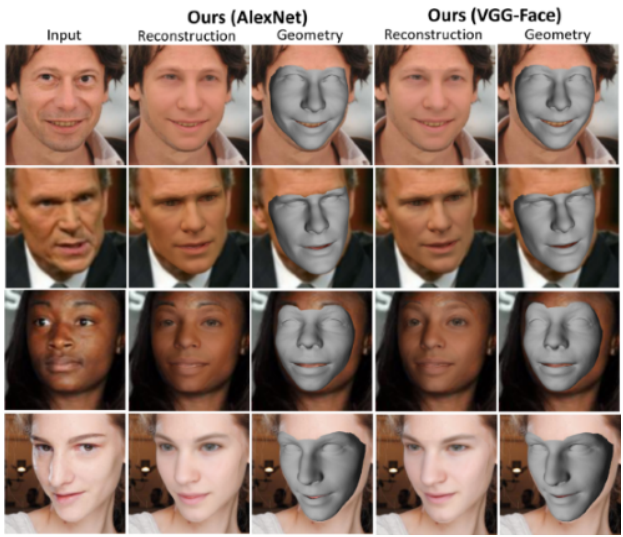

Figure 6. We evaluate different encoders in combination with our model-based decoder. In average VGG-Face [7] leads to slightly better results than AlexNet [5], but the results are comparable.

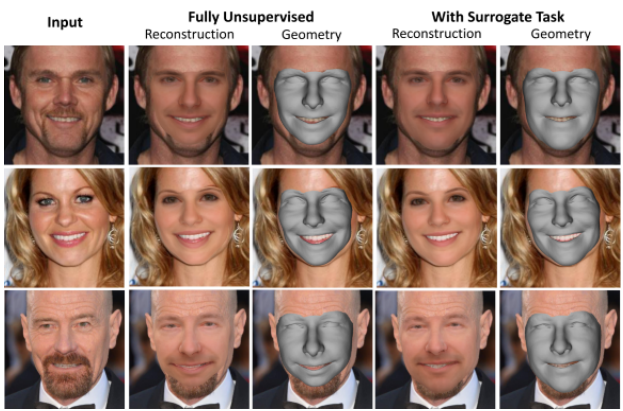

Figure 7. We evaluate the influence of the proposed surrogate task. The surrogate task leads to improved reconstruction quality and increases robustness to occlusions and strong expressions.

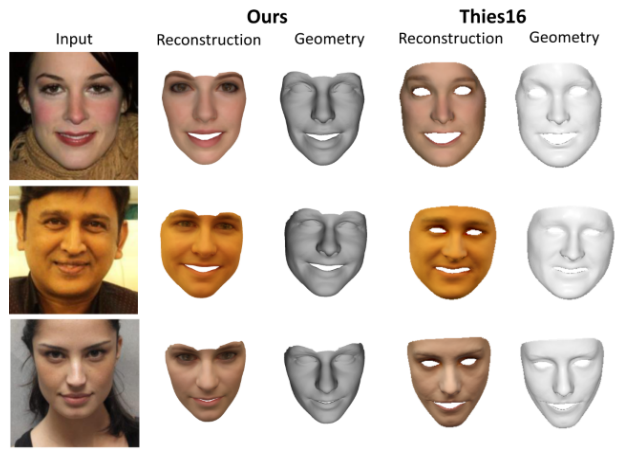

Figure 10. Comparison to the monocular reconstruction approach of [12] on CelebA [6]. Our approach obtains similar or higher quality, while being orders of magnitude faster (4ms vs. $\sim$ 500ms).